\pdfoutput=1

\documentclass[11pt]{article}

\usepackage{naacl2021}

\usepackage{times}
\usepackage{latexsym}

\usepackage[T1]{fontenc}
\usepackage{xcolor}
\mathchardef\mhyphen="2D 
\usepackage{graphicx}
\usepackage{multirow}
\usepackage{amsmath}
\usepackage{tabularx}

\usepackage[utf8]{inputenc}

\usepackage{microtype}

\title{Seed Word Selection for Weakly-Supervised Text Classification \\ with Unsupervised Error Estimation}

\author{Yiping Jin$^{1,2}$, Akshay Bhatia$^2$, Dittaya Wanvarie$^1$ \\
$^1$Department of Mathematics \& Computer Science, Chulalongkorn University, Thailand\\
$^2$Knorex, 140 Robinson Road, 14-16 Crown @ Robinson, Singapore\\
\texttt{\{jinyiping, akshay.bhatia\}@knorex.com} \\
\texttt{Dittaya.W@chula.ac.th} \\
}

\begin{document}
\maketitle
\begin{abstract}
Weakly-supervised text classification aims to induce text classifiers from only a few user-provided seed words. The vast majority of previous work assumes high-quality seed words are given. However, the expert-annotated seed words are sometimes non-trivial to come up with. Furthermore, in the weakly-supervised learning setting, we do not have any labeled document to measure the seed words' efficacy, making the seed word selection process ``a walk in the dark''. In this work, we remove the need for expert-curated seed words by first mining (noisy) candidate seed words associated with the category names. We then train interim models with individual candidate seed words. Lastly, we estimate the interim models' error rate in an unsupervised manner. The seed words that yield the lowest estimated error rates are added to the final seed word set. A comprehensive evaluation of six binary classification tasks on four popular datasets demonstrates that the proposed method outperforms a baseline using only category name seed words and obtained comparable performance as a counterpart using expert-annotated seed words~\footnote{Source code can be found at \url{https://github.com/YipingNUS/OptimSeed}.}.
\end{abstract}

\section{Introduction}
\label{sec:intro}

Weakly-supervised text classification eliminates the need for any labeled document and induces classifiers with only a handful of carefully chosen seed words. However, some researchers pointed out that the choice of seed words has a significant impact on the performance of weakly-supervised models~\cite{li2018seed,jin1learning}. The vast majority of previous work assumed high-quality seed words are given. However, many seed words reported in previous work are not intuitive to come up with. For example, in Meng et al.~\shortcite{meng2019weakly}, the seed words used for the category ``Soccer'' are \{cup, champions, united\} instead of more intuitive keywords like ``soccer'' or ``football''. We conjecture the authors might have tried these more general keywords but avoided them because they do not perform well.

While it is common to use labeled corpora to evaluate weakly-supervised text classifiers in the literature, we do not have access to any labeled document for new categories in the real-world setting. Therefore, there is no way to measure the model's performance and select the seed words that yield the best accuracy. A similar concern on assessing active learning performance at runtime has been raised by Kottke et al. ~\shortcite{kottke2019limitations}.

In this work, we device $OptimSeed$, a novel framework to automatically compose and select seed words for weakly-supervised text classification. We firstly mine (noisy) candidate seed words associated with the category names. We then train interim models with individual candidate seed words in an iterative manner. Lastly, we use an unsupervised error estimation method to estimate the interim models' error rates. The keywords that yield the lowest estimated error rates are selected as the final seed word set. A comprehensive evaluation of six classification tasks on four popular datasets demonstrates the effectiveness of the proposed method. The proposed method outperforms a baseline using only the category name as seed word and obtained comparable performance as a counterpart using expert-annotated seed words. We use binary classification as a case study in this work, while the idea can be generalized to multi-class classification using one-vs.-rest strategy.

The contributions of this work are three-fold:
\begin{enumerate}
	\itemsep0em
	\item We propose a novel combination of unsupervised error estimation and weakly-supervised text classification to improve the classification performance and robustness.
	\item We conduct an in-depth study on the impact of different seed words on weakly-supervised text classification, supported by experiments with various models and classification tasks.
	\item The proposed method generates keyword sets that yield consistent and competitive performance against expert-curated seed words.
\end{enumerate}

\section{Related Work}
\label{sec:relatedworks}

We review the literature in three related fields: (1) weakly-supervised text classification, (2) unsupervised error estimation, and (3) keyword mining.

\subsection{Weakly-Supervised Text Classification}
\label{subsec:relatedworks:weakly-supervised}

Weakly-supervised text classification~\cite{druck2008learning,meng2018weakly,meng2019weakly} aims to use a handful of labeled seed words to induce text classifiers instead of relying on labeled documents.

Druck et al.~\shortcite{druck2008learning} proposed generalized expectation (GE), which specifies the expected posterior probability of labeled seed words appearing in each category. GE is trained by optimizing towards satisfying the posterior constraints without making use of pseudo-labeled documents.

Chang et al.~\shortcite{chang2008importance} introduced the first embedding based weakly-supervised text classification method. They mapped category names and documents into the same semantic space. Document classification is then performed by searching for the nearest category embedding given an input document. 

Meng et al.~\shortcite{meng2018weakly} proposed weakly-supervised neural text classification. They generate unambiguous pseudo-documents, which are used to induce text classifiers with different architectures such as convolutional neural networks~\cite{kim2014convolutional} or Hierarchical Attention Network~\cite{yang2016hierarchical}.

Recently, Mekala and Shang~\cite{mekala2020contextualized} disambiguate the seed words by explicitly learning different senses of each word with contextualized word embeddings. They first performed k-means clustering for each word in the vocabulary to identify potentially different senses, then eliminated the ambiguous keyword senses.

Two most recent works developed concurrently but independently from our work~\cite{meng2020text,wang2020xclass} addressed the same task we are tackling: weakly-supervised text classification from only the category name. They both tap on the category names' contextualized representation and expand the seed word list by finding other words that would fit into the same context.

\subsection{Unsupervised Error Estimation}
\label{subsec:relatedworks:ee}

Unsupervised error estimation aims to estimate the error rate of a list of classifiers \textit{without a labeled evaluation dataset}. It is widely relevant to machine learning models in production, such as when a pre-trained model is applied to a new domain or when labeled dataset is costly to obtain. To our best knowledge, no previous work in weakly-supervised classification applied unsupervised error estimation. Instead, they trained classifiers without labeled \textit{training} datasets but evaluated their models used labeled \textit{evaluation} datasets.

Most work in unsupervised error estimation derive the error rate analytically by making simplifying assumptions. Donmez et al.~\shortcite{donmez2010unsupervised} and Jaffe et al.~\shortcite{jaffe2015estimating} assumed the marginal probability of the category $p(y)$ is known. Platanios et al.~\shortcite{10.5555/3020751.3020822} assumed classifiers make conditionally independent errors. While these approaches laid an important theoretical foundation, most assumptions cannot be met for real-world datasets and classifiers. 

Platanios et al.~\shortcite{platanios2016estimating} proposed a Bayesian approach for error estimation. The model infers the true category and the error rates jointly using Gibbs sampling. The approach was benchmarked with various baselines such as majority vote and Platanios et al.~\shortcite{10.5555/3020751.3020822} and achieved superior performance. The estimated accuracy is usually within a few percents from the true accuracy. Notably, the only mild assumption it makes is that more than half of the classifiers have an error rate lower than 50\%. 


\subsection{Keyword Mining}
\label{subsec:relatedworks:keyword Mining}

Keyword mining aims to bootstrap high-quality keyword lexicons from a small set of seed words, and it has been widely used in mining opinion lexicons~\cite{hu2004mining,hai2012one} and technical glossaries~\cite{elhadad-sutaria-2007-mining}. We want to draw the association between keyword mining and weakly-supervised text classification. Both tasks take a small list of seed words and unlabeled corpus as input, aiming to ``expand'' the knowledge about the target semantic category. Having more high-quality keywords will improve classification accuracy, while an accurate classifier will make the keyword mining task much easier by eliminating irrelevant/noisy documents.

\section{Method}
\label{sec:method}

\begin{figure*}
  \centering \includegraphics[width=465pt]{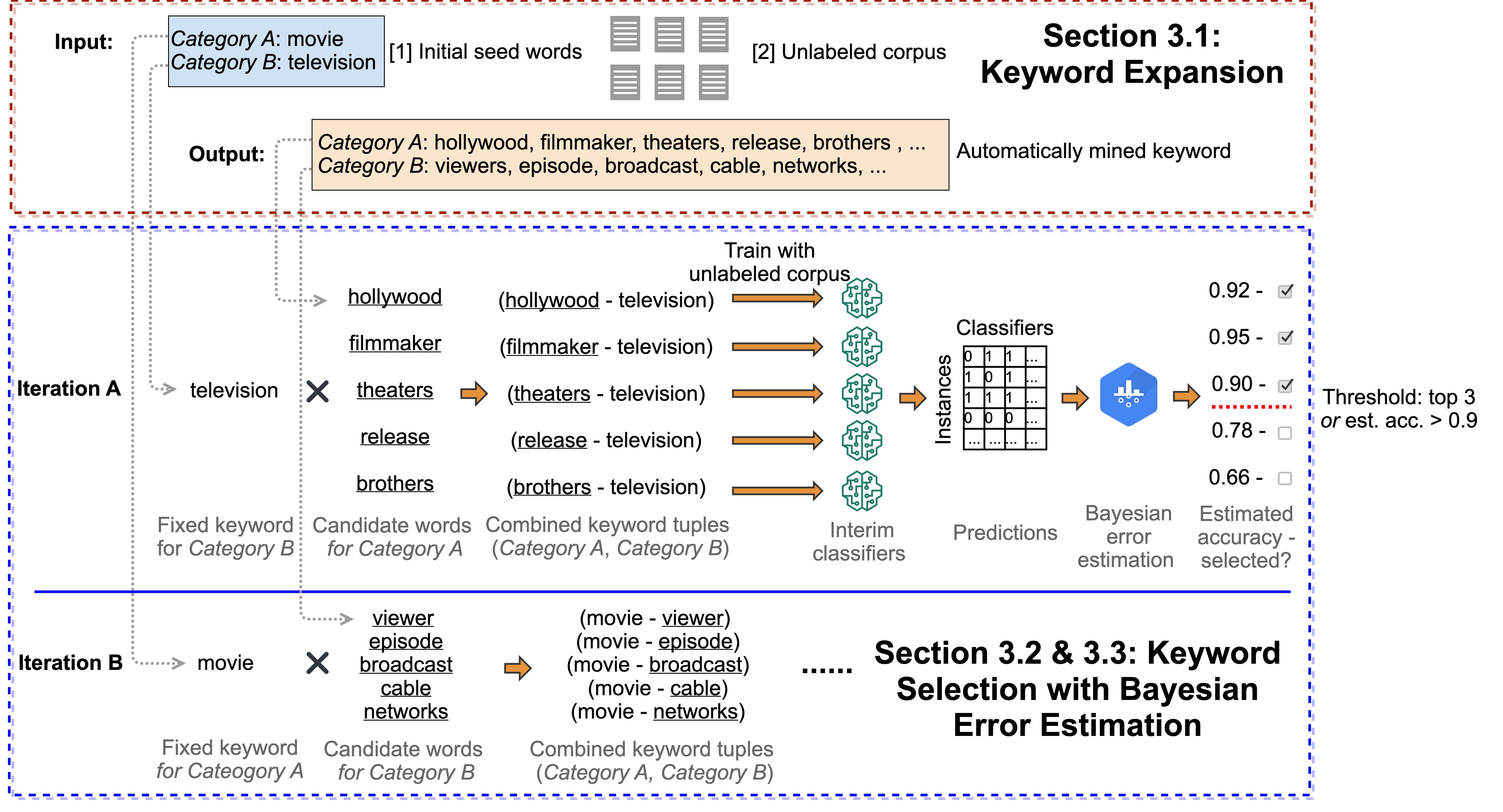}
  \caption{OptimSeed, a method to select seed words for weakly-supervised text classification. We first mine noisy keywords associated with the category name (the initial seed word). We use one iteration to refine the keywords for each category. In each iteration, We fix the seed word for one category and combine it with each mined keyword in the other category. The resultant keyword tuples are used to train separate interim classifiers. Finally, we use Bayesian error estimation to estimate the accuracy of classifiers induced from each keyword tuple and select the keywords with the highest estimated accuracy.}
  \label{fig:overview}
\end{figure*}

Figure~\ref{fig:overview} overviews OptimSeed, a framework to select seed words for weakly-supervised text classification involving the following steps: (1) expanding candidate keywords from a single seed word, (2) training interim classifiers with individual candidate seed keywords using weakly supervision, (3) select the final seed words with the feedback from unsupervised error estimation. We discuss the proposed framework in detail in the following sections. To make our paper self-contained, we also brief the weakly-supervised classification and unsupervised error estimation model we use.

\subsection{Expanding Candidate Keywords from a Single Seed}
\label{subsec:method:kw-expansion}

We use either the category name or trivial keywords (e.g., ``good'' and ``bad'' for sentiment classification tasks) as the only input seed word and use a keyword expansion algorithm to mine more candidate keywords. We apply $pmi$-$freq$ (Equation~\ref{eq:pmi-freq}) following Jin et al.~\shortcite{jin1learning}. It is a product of the logarithm of the candidate keyword $w$'s document frequency and the point-wise mutual information between $w$ and the seed word $s$. The higher the $pmi$-$freq$ score, the more strongly the candidate keyword is associated with the seed word $s$. Additionally, we filter the mined keywords based on their part-of-speech tag depending on the classification task. We keep only noun candidates for topic classification and adjective candidates for sentiment classification. 

\begin{equation} \label{eq:pmi-freq}
pmi\mhyphen freq(w;s) \equiv log\,df(w) log \frac{p(w,s)}{p(w)p(s)}
\end{equation}

\subsection{Training interim classifiers}
\label{subsec:method:interim-model}

The candidate keywords and unlabeled dataset are used to induce \textit{interim classifiers}. Interim classifiers' purpose is to isolate the impact of individual seed words so that we can rank them. Specifically, iteration A in Figure~\ref{fig:overview} tries to rank candidate seed words for Category A (Movies) in the classification task Movies-Television. The initial seed word ``television'' for Category B is fixed, and it forms seed word tuples with each candidate word in Category A. We use each such seed word tuple as input to train an interim classifier. We then use each interim classifier's predictions to perform unsupervised error estimation.

We use Generalized Expectation (GE)~\cite{druck2008learning} to train both interim classifiers and the final classifier because of its competitive performance and fast training speed~\footnote{All GE models in this work can be trained within a few seconds using a single CPU core.}. GE translates labeled keywords to constraint functions. For example, the first keyword tuple (hollywood, television) in Figure~\ref{fig:overview} translates to two constraint functions:  $hollywood\rightarrow{A: 0.9, B: 0.1}$ and $television\rightarrow{A: 0.1, B: 0.9}$, which means ``hollywood'' is expected to occur 90\% in a document of category A while 10\% in a document of category B, vice versa for the keyword ``television''. 

Each constraint function on a labeled word $w_k$ contributes to a term in the objective function in Equation~\ref{eq:ge} and the underlying logistic regression model is trained by minimizing the L2 distance between the reference distribution $\hat{p}(y|w_k>0)$ (specified by the constraint function) and the empirical distribution $\tilde{p}(y|w_k>0)$ (predicted by the model) of the category $y$ when word $w_k$ is present.

\begin{equation} \label{eq:ge}
\mathcal O= -\sum_{k \in K}dist(\hat{p}(y|w_k>0)||\tilde{p}(y|w_k>0))
\end{equation}

\subsection{Keyword Evaluation with Bayesian Error Estimation}
\label{subsec:method:bee}

We apply unsupervised error estimation on the interim classifiers' predictions to estimate their accuracy and select the best seed words for the final classifier. As shown in Figure~\ref{fig:overview} iteration A, the three keywords ``hollywood'', ``filmmaker'', and ``theaters'' are added to the final seed word set of Category A (Movies) because their corresponding interim classifiers have estimated accuracy above the threshold. The process is repeated in iteration B to select seed words for Category B.

We use the Bayesian error estimation (BEE) model~\cite{platanios2016estimating} to perform this step. In BEE, each instance's true label is latent, while each model's predictions are observed. The accuracy/error rate can be derived from the predictions and the latent true labels. The assumption that half of the classifiers have an error rate below 50\% implicitly uses inter-classifier agreement. 

BEE uses Gibbs sampling to infer the error rates of individual classifiers $e_j$ and the true label $l_i$ jointly. We refer the readers to Section 4.1 in Platanios et al.~\shortcite{platanios2016estimating} for the exact conditional probabilities used in Gibbs sampling.

\section{Experimental Setup}
\label{sec:experiment}

We use six binary classification tasks from four datasets to evaluate our framework. We choose the evaluation tasks so that they cover different granularities and domains. The details are as follows:

\begin{itemize}
	\itemsep0em
	\item \textbf{AG's News Dataset:} contains 120,000 documents evenly distributed into 4 coarse categories. We randomly choose two binary classification tasks: ``Politics'' vs. ``Technology'' and ``Business'' vs. ``Sports''.
	\item \textbf{The New York Times (NYT) Dataset:} contains 13,081 news articles covering 5 coarse and 25 fine-grained categories. We choose two fine-grained binary classification tasks involving categories with similar semantics: ``International Business'' (InterBiz) vs. ``Economy'' and ``Movies'' vs. ``Television''.
	\item \textbf{Yelp Restaurant Review Dataset:} contains 38,000 reviews evenly distributed into 2 categories: ``Positive'' vs. ``Negative''.
	\item \textbf{IMDB Movie Review Dataset:} contains 50,000 reviews evenly distributed into 2 categories: ``Positive'' vs. ``Negative''.
\end{itemize}

We report the performance of the following weakly-supervised models besides Generalized Expectation (GE):

\begin{itemize}
	\itemsep0em
	\item \textbf{Dataless~\cite{chang2008importance}:} maps both input documents and category seed words into a semantic space using Explicit Semantic Analysis (ESA)~\cite{gabrilovich2007computing} over Wikipedia concepts and assigns the category nearest to the input document's embedding. 
	\item \textbf{MNB/Priors~\cite{settles2011closing}:} increases priors for labeled keywords in a Na\"ive Bayes model and learns from an unlabeled corpus using EM algorithm.
	\item \textbf{\textsc{W{\scriptsize E}STC{\scriptsize LASS}}~\cite{meng2018weakly}:} weakly-supervised neural text classifier trained using pseudo documents. We use the CNN architecture because Meng et al.~\shortcite{meng2018weakly} showed that it outperformed other architectures such as RNNs and Hierarchical Attention Network.
	\item \textbf{ConWea~\cite{mekala2020contextualized}:} leverages contextualized word representations to differentiate multiple senses. It also trains classifiers and expands seed words in an iterative manner. 
\end{itemize}

We also report the performance of \textbf{LR}, a supervised logistic regression model trained using all the documents in the training set~\footnote{We use the logistic regression implementation in scikit-learn with default parameters and tf-idf features.}. 

In all experiments, we mine 16 candidate seed words with the highest $pmi$-$freq$ score for each category. We select a candidate keyword for the final classifier if its estimated accuracy is among the top three or is higher than 0.9~\footnote{Mekala and Shang~\shortcite{mekala2020contextualized} observed that three seed words per class are needed for reasonable performance while more high-quality keywords help. Therefore, we use the accuracy threshold of 0.9 to include additional keywords.}. For GE, we use a reference distribution of 0.9 (meaning a labeled keyword is expected to appear in its specified categories 90\% of the time) following Druck et al.~\shortcite{druck2008learning}. Table~\ref{tab:seedwords} shows the initial seed words used in our work and in previous work~\footnote{The seed words for NYT corpus were reported in Meng et al.~\shortcite{meng2019weakly} and the rest are from Meng et al.~\shortcite{meng2018weakly}. No previous work in weakly supervision used IMDB dataset, so we use the same manual seed words as Yelp dataset.}. 

\begin{table}[!htbp]
\centering
\begin{tabularx}{\textwidth}{p{1.5cm}p{1.8cm}p{3cm}}
\cline{1-3}
\textbf{Class} & \textbf{Our Work} & \textbf{Previous Work} \\ \cline{1-3}
Politics \newline\newline Tech & political; \newline\newline technology & democracy religion liberal; \newline scientists biological computing  \\ \cline{1-3}
Business \newline\newline Sports & business; \newline\newline sports & economy industry \newline investment; \newline hockey tennis basketball  \\ \cline{1-3}
InterBiz \newline Economy & international; \newline economy & china union euro; \newline fed economists economist  \\ \cline{1-3}
Movies \newline\newline Television & movie; \newline\newline television & hollywood directed oscar; \newline episode viewers episodes \\ \cline{1-3}
Yelp \& IMDB & good; \newline\newline bad &  terrific great \newline awesome; \newline horrible subpar\newline disappointing \\  \cline{1-3}
\end{tabularx}
\caption{Initial seed words for each task.}
\label{tab:seedwords}
\end{table}

\section{Classification Performance}
\label{sec:experiment:classification}

Table~\ref{tab:classification-avg-acc} presents each model's average accuracy across six datasets.

\begin{table}[!htbp]
\centering
  \begin{tabular}{p{2.0cm}p{1cm}p{1cm}p{1cm}}
    \hline
    {\textbf{Method}} & \textbf{cate} & \textbf{ours} & \textbf{gold} \\
    \hline
    Dataless & 54.7 & \textbf{60.4}$^*$ & 56.7  \\ 
    MNB/Priors & 68.5 & 71.7 & \textbf{74.4} \\ 
    \textsc{W{\scriptsize E}STC{\scriptsize LASS}} & 75.7 & \textbf{77.2} & 77.0 \\ 
    ConWea & 60.0 & 66.0 & \textbf{70.7} \\ 
    \hline
    GE & 80.4 & 84.8$^*$ & \textbf{85.1} \\
    LR & \multicolumn{3}{c}{91.8}  \\  \hline
  \end{tabular}
  
\caption{Average accuracy scores in percentage for all methods on all six classification tasks. \textbf{cate}, \textbf{ours}, \textbf{gold} indicates the result
using the category name, keywords selected by OptimSeed and expert-composed keywords used in previous work. For each model, the best-performing keyword set is highlighted in bold. $^*$ indicates statistical significance from the same model using ``cate'' seed word with p-value of 0.05 using paired t-test.}
\label{tab:classification-avg-acc}
\end{table}

We can see that OptimSeed seed words yield better performance than using the category name alone by a large margin for all weakly-supervised models, validating the effectiveness of our seed word expansion and selection method. It also achieved better or similar performance as expert-curated seed words for three out of five models. 

Among the learning algorithms, GE obtained the best average performance for all seed word sets. The average accuracy of GE using OptimSeed seed words (84.8\%) is only 0.3\% lower than using expert-curated seed words, virtually eliminating human experts from the loop. GE+OptimSeed's accuracy is 7\% below a fully-supervised logistic regression model trained on hundreds to tens of thousands of labeled documents.

Table~\ref{tab:classification-acc} shows each model's classification accuracy on topic classification tasks. Summing over all models and datasets, OptimSeed achieved better or equal performance than the category name baseline 80\% of the time (16/20) and better or equal performance than the gold seed words 65\% of the time. It demonstrates the robustness of our seed word selection method across different tasks.

While ConWea claimed to resolve ambiguity through contextualized embeddings, we observed that it works well only when the input seed words are unambiguous (``ours'' or ``gold'' column). On the Business-Sports classification task, its accuracy was only 39.1\% while other baselines could achieve over 90\%. We inspected the model and found the keywords expanded by ConWea are much noisier than OptimSeed, which caused the poor performance.

\begin{table*}[!htbp]
\centering
  \begin{tabular}{p{2.0cm}p{0.55cm}p{0.55cm}p{0.7cm}p{0.55cm}p{0.55cm}p{0.7cm}p{0.55cm}p{0.55cm}p{0.7cm}p{0.55cm}p{0.55cm}p{0.55cm}}
    \hline
    \multirow{2}{*}{\textbf{Method}} &
      \multicolumn{3}{c}{\textbf{Poli-Tech}} &
      \multicolumn{3}{c}{\textbf{Biz-Sport}} &
      \multicolumn{3}{c}{\textbf{IB-Econ}} & 
      \multicolumn{3}{c}{\textbf{Movie-TV}} \\ 
    & \textbf{cate} & \textbf{ours} & \textbf{gold} & \textbf{cate} &  \textbf{ours} & \textbf{gold} & \textbf{cate} & \textbf{ours} & \textbf{gold} & \textbf{cate} & \textbf{ours} & \textbf{gold}\\
    \hline
    Dataless & 50.1 & \textbf{51.4} & 50.2 & 50.0 & 50.2 & \textbf{50.4} & 59.1 & \textbf{75.0} & 67.1 & 67.8 & \textbf{70.0} & 67.8\\ 
    MNB/Priors & 87.3 & \textbf{88.9} & \textbf{88.9} & \textbf{95.6} & 93.9 & 92.9 & 58.5 & 54.3 & \textbf{93.9} & 67.8 & 67.8 & \textbf{68.9} \\ 
    \textsc{W{\scriptsize E}STC{\scriptsize LASS}} & 87.4 & \textbf{89.5} & 88.8 & 92.7 & \textbf{94.8} & 94.3 & 77.7 & \textbf{83.0} & 75.1 & 50.4 & \textbf{76.6} & 62.1 \\ 
    ConWea & 71.5 & \textbf{73.7} & 71.4 & 39.1 & 67.0 & \textbf{82.0} & 75.1 & 71.2 & \textbf{84.3} & 66.9 & \textbf{77.0} & 76.4 \\ 
    \hline
    GE & 86.9 & 87.8 & \textbf{88.5} & \textbf{93.0} & \textbf{93.0} & 79.4 & 70.7 & 81.7 & \textbf{91.5} & 94.4 & \textbf{98.9} & 97.8 \\
    LR & \multicolumn{3}{c}{96.3} & \multicolumn{3}{c}{98.6} & \multicolumn{3}{c}{90.2} & \multicolumn{3}{c}{85.5} \\  \hline
  \end{tabular}
  
\caption{Accuracy on topic classification tasks. For each model-dataset combination, we highlight the best performance in bold.}
\label{tab:classification-acc}
\end{table*}

We can make similar observations on the performance of sentiment classification tasks (Table~\ref{tab:sentiment-classification-acc}). However, the gap between weakly-supervised models and the supervised baseline is much larger topic classification tasks, suggesting that some reviews' sentiment might be expressed implicitly and requires more than word-level understanding. \citet{meng2020text} also made a similar remark based on their experiment. 

\begin{table}[!htbp]
\centering
   \begin{tabular}{p{2.0cm}p{1cm}p{1cm}p{1cm}}
    \hline
    \multirow{2}{*}{\textbf{Method}} &
      \multicolumn{3}{c}{\textbf{Yelp}}\\ 
    & \textbf{cate} & \textbf{ours} & \textbf{gold}\\
    \hline
    Dataless & 51.0 & \textbf{55.5} & 52.2 \\ 
    MNB/Priors & 50.9 & \textbf{71.5} & 51.7 \\ 
    \textsc{W{\scriptsize E}STC{\scriptsize LASS}} & 78.3 & 58.8 & \textbf{81.5} \\ 
    ConWea & 51.0 & \textbf{51.3} & 50.7 \\ 
    \hline
    GE & 68.0 & 75.2 & \textbf{79.3} \\
    LR & \multicolumn{3}{c}{92.2} \\  \hline
  \end{tabular}
 
   \begin{tabular}{p{2.0cm}p{1cm}p{1cm}p{1cm}}
    \multirow{2}{*}{\textbf{Method}} &
      \multicolumn{3}{c}{\textbf{IMDB}}\\ 
    & \textbf{cate} & \textbf{ours} & \textbf{gold}\\
    \hline
    Dataless & 50.1 & \textbf{60.4} & 52.2 \\ 
    MNB/Priors & 51.1 & \textbf{54.0} & 50.3 \\ 
    \textsc{W{\scriptsize E}STC{\scriptsize LASS}} & \textbf{67.7} & 60.6 & 60.5 \\ 
    ConWea & 56.5 & 55.7 & \textbf{59.1} \\ 
    \hline
    GE & 69.6 & 72.2 & \textbf{74.0} \\
    LR & \multicolumn{3}{c}{88.3} \\  \hline
  \end{tabular}
   
\caption{Accuracy on sentiment classification tasks. For each model-dataset combination, we highlight the best performance in bold.}
\label{tab:sentiment-classification-acc}
\end{table}

\section{Case Study}
\label{sec:experiment:case-study}

To demonstrate the working of our proposed framework, we present a case study  on the classification task ``International Business'' vs. ``Economy'' in Table~\ref{tab:case-study} and show different seed word sets for the category ``economy'' and their corresponding performance. 

Keyword expansion alone improved the accuracy significantly from the category name baseline. However, it may introduce some ambiguous keywords in the meantime. The unsupervised error estimator successfully identified top keywords such as ``economist'' and ``economists'' and eliminated poor keywords like ``purchases'' and ``growth'', which further improved the accuracy by 2.4\%.

\begin{table}[!htbp]
\centering
  \begin{tabularx}{\textwidth}{p{1.8cm}p{5cm}}
    \cline{1-2}
    \textbf{Stage:Acc} & \textbf{Seed Words for ``Economy''}\\
    \cline{1-2}
    Init: 70.7 & economy   \\
    \cline{1-2}
    Keyword Expansion: 79.3 & purchases pace index \newline borrowing unemployment \newline economists economy stimulus \newline rates recovery economist rate \newline fed reserve inflation growth \\ \cline{1-2}
    Final: 81.7 & economist economists rate\newline recovery index \\ \cline{1-2}
  \end{tabularx}
  
\caption{Seed words for ``Economy'' at different stages of the OptimSeed framework.}
\label{tab:case-study}
\end{table}

\section{Conclusion}
\label{sec:conclusion}

Weakly-supervised text classification can induce classifiers with a handful of carefully-chosen seed words instead of labeled documents. However, the choice of seed words has a significant impact on classification performance. We proposed $OptimSeed$, a novel framework to compose the seed words automatically. It first mines keywords associated with the category name and then estimates each seed word's impact directly using unsupervised error estimation. The framework outputs seed words yielding a comparable performance as expert-curated seed words, virtually eliminating human experts from the loop.


\section{Acknowledgements}

YJ was supported by the scholarship from `The 100th Anniversary Chulalongkorn
University Fund for Doctoral Scholarship'. We thank anonymous reviewers for their valuable feedback.

\bibliography{custom}
\bibliographystyle{acl_natbib}

\end{document}